\documentclass[journal]{IEEEtran}

\usepackage[pdftex]{graphicx}
\graphicspath{{figures/}}
\usepackage[cmex10]{amsmath}
\usepackage{array}
\usepackage[tight,footnotesize]{subfigure}
\usepackage{url}
\usepackage{verbatim}

\begin{document}
%
\title{Improved State Estimation in Quadrotor MAVs: A Novel Drift-Free Velocity Estimator}

\author{Dinuka~Abeywardena,
        Sarath~Kodagoda,
        Gamini~Dissanayake,
		  and~Rohan~Munasinghe
\thanks{D. Abeywardena S. Kodagoda and G. Dissanayake are with the Centre for Autonomous Systems, University of Technology Sydney,
NSW, 2007 Australia. R. Munasinghe is with the Department of Electronic and Telecommunication Engineering, University of Moratuwa, Sri Lanka }}

\maketitle

\begin{abstract}
This paper describes the synthesis and evaluation of a novel state estimator for a Quadrotor Micro Aerial Vehicle.  Dynamic equations which relate acceleration, attitude and the aero-dynamic propeller drag are encapsulated  in an extended Kalman filter framework for estimating the velocity and the attitude of the quadrotor.  It is demonstrated that exploiting the relationship between the body frame accelerations and velocities, due to blade flapping, enables  drift free estimation of lateral and longitudinal components of body frame translational velocity along with improvements to roll and pitch components of body attitude estimations. Real world data sets gathered using a commercial off-the-shelf quadrotor platform, together with ground truth data from a Vicon system, are used to evaluate the effectiveness of the proposed algorithm.
\end{abstract}

\section*{Introduction}
Quadrotor Micro Aerial Vehicles (MAV) are simple robotic platforms to construct. In its' basic form, it is no more than two counter rotating propeller pairs attached symmetrically to a rigid cross-like frame, along with the means to control the speed of each individual propeller. This symmetric design is what has enabled the quadrotor to become a simple yet powerful vertical take-off and landing aerial platform that is popular among the robotics community.

With this simplicity comes the burden of controlling motion in 3D space with the use of just four actuators. Under-actuated and coupled dynamics of the quadrotor make it nearly impossible for a human pilot to gain control of it, unless a well tuned control system is in place. Such a control system is also vital if autonomy is a goal, as is the case with most MAVs. Estimates of controlled states and their derivatives are essential for any control system, and where those estimates are accurate and frequent in time, it has been demonstrated that quadrotors have extreme maneuverability and agility\cite{mellinger2010}. 

However, MAVs are - by design - limited in their payload capacity and with those limitation, obtaining accurate \textit{and} fast state estimates becomes a challenge. For example, MEMS inertial sensors can provide fast but coarse state estimates \cite{king2004}, while exteroceptive sensors such as lasers and cameras\cite{markus2008} render more accurate state estimates albeit at a slower rate. Attempts to merge these two sensing domains are frequent in MAV literature \cite{bryson2007}, \cite{taylor2009} and an application of similar ideas to quadrotors was presented in \cite{ahrens2009}. 

One aspect common to most MAV state estimators is  their use of inertial sensors. Typically gyroscopes, accelerometers and magnetometers are used for the purpose of attitude estimation \cite{flux2008}.  Based on a long history of research in inertial navigation systems, sensor fusion algorithms usually employed for this task make use of the equations of motion of the sensing unit in three-dimensional space.  The main advantage of this approach is that these \textit{generic} estimators are specific only to the sensor package geometry and as such can be used independently of the platform on which the sensors are mounted. However, they fail to exploit the dynamics of the vehicle under consideration in the estimation process, leading to a potentially sub-optimal result. The value of using specific dynamic characteristics of the vehicle has been reported in the case of land vehicles \cite{dissa1999} and air vehicles\cite{tahk1990}.  Similarly, in this paper we demonstrate that the influence of blade flapping in a quadrotor leads to a set of dynamic equations that can aid state estimation using inertial sensors.  

The rest of the paper is arranged as follows. First, some background on quadrotor state estimators and a discussion on what motivated us to look beyond the state-of-the-art is presented. We will then briefly present the quadrotor dynamic equations that are of interest to the state estimation process. After highlighting the shortcomings of the generic design, a novel state estimator design is presented along with experimental results which demonstrate the accuracy and consistency of estimates. The article is conclude by exploring the implications of the novel algorithm.

\section*{Background and Motivation}
MAV attitude estimators that fuse gyroscope and accelerometer measurement using generic algorithms are frequently reported in literature \cite{kumar2004}, \cite{flux2008}, \cite{king2004}. In a nutshell, these algorithms operate by fusing measurements of a triad of body mounted gyroscopes and accelerometers. Gyroscope measurements are a source of high frequency attitude rate information, but they alone are not sufficient for  drift free attitude estimation due to bias and various other forms of noise present in a typical low cost sensor. Attitude estimators for MAVs overcome this issue by assuming that  accelerometers predominantly measure gravitational acceleration and are thus capable of providing low frequency information about MAV orientation with respect to gravity. Clearly, when the vehicle accelerations are significant,  as in the case of  quadrotor, this assumption does not hold \cite{dmw2010}.  Furthermore, such estimators  are incapable of drift free velocity estimation, as they can only be generated by integrating noisy accelerometer measurements. To complicate the matters even further, accelerometer measurements need to be compensated for gravity before this integration, and such a compensation requires an accurate attitude estimate. As mentioned before, one promising way to overcome these deficiencies is to examine the behaviour of the MAV in question, in-order to identify suitable characteristics that would assist the estimation process. 

Martin et. al. \cite{martin2009} have analysed the behaviour of a quadrotor MAV in detail and also presented equations describing measurements of an accelerometer mounted on-board a quadrotor. Their results motivated us to reformulate the state estimators for quadrotors and to redesign them considering the true sensor behaviour as opposed to conventional vehicle independent assumptions. In addition to improving the accuracy of the attitude estimate, the design presented here provides a drift free estimate of the horizontal components of translational velocity of the quadrotor. Recently, a similar idea was presented in \cite{mahony2012} where two separate non-linear complementary filters were utilised to estimate attitude and velocity of a quadrotor MAV. The filter formulation presented in this paper is different from \cite{mahony2012} and we also present experimental results validating the concept. The velocity estimates thus derived are of critical importance to control and navigational tasks of a quadrotor, as will be discussed in concluding remarks.

\section*{Quadrotors: What Makes Them Unique?}
A thorough derivation and analysis of the quadrotor dynamics can be found in \cite{bris2009} and \cite{martin2009}. Rather than re-iterating the derivation, here we aim to briefly summarise the important equations and to provide an intuitive description of the most salient features of the dynamic behaviour that makes quadrotors a unique MAV.

\begin{figure}[htb]
\begin{center}
\includegraphics[width=0.5\textwidth]{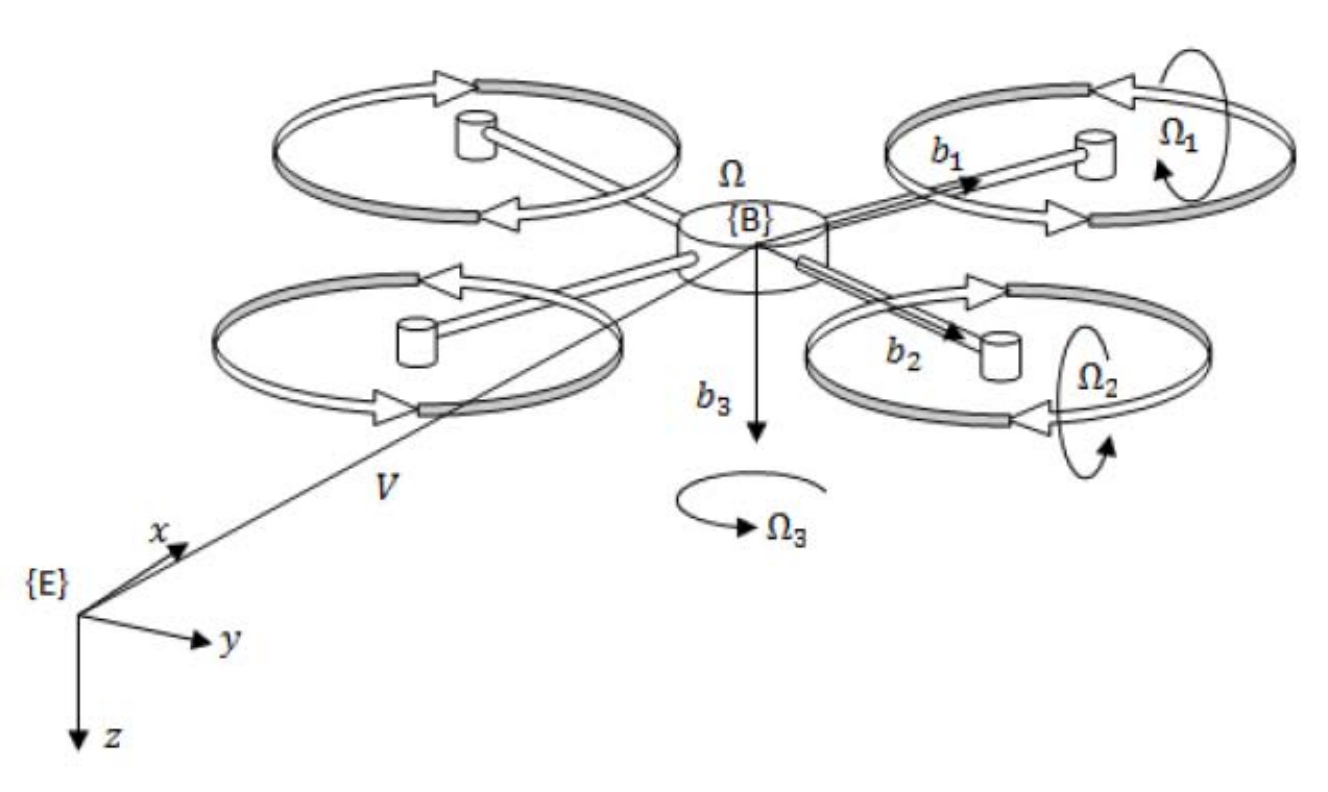}
\end{center}
\caption{Coordinate frame definitions for the quadrotor dynamic model}
\label{fig_frame_def}
\end{figure}

Let $\{E\}$ be the earth fixed inertial frame, and a vector $[\begin{array}{ccc}x&y&z\end{array}]^T$ denote the position of the centre of mass of the quadrotor expressed in $\{E\}$ (See Fig. \ref{fig_frame_def}). Let $\{B\}\equiv[\begin{array}{ccc}b_1&b_2&b_3 \end{array}]^T $ be a body fixed frame positioned at the centre of mass of the quadrotor.

The orientation of $\{B\}$ with respect to $\{E\}$ is defined using a cumulative rotation of Euler angles $\psi$ (Yaw) , $\theta$ (Pitch) and $\phi$ (Roll) in that order, around $b_3$, $b_2$ and $b_1$, respectively. ${R}$ is defined as the rotational transformation matrix from $\{B\}$ to $\{E\}$.
The kinematic equation relating the instantaneous angular velocity $\textbf{$\Omega$}\equiv[\begin{array}{ccc}\omega_x&\omega_y&\omega_z\end{array}]$ of  $\{B\}$ with respect to $\{E\}$, to Euler rates can be expressed as:

\begin{equation}\label{eq_rotkin}
\begin{bmatrix} \dot{\phi} \\ \dot{\theta}  \\ \dot{\psi} \end{bmatrix} = 
\begin{bmatrix} 1 & \tan\theta \sin\phi & \tan\theta \cos\phi \\
0 & \cos\phi & -\sin\phi \\
0 & \sin\phi / \cos\theta & \cos\phi / \cos\theta
\end{bmatrix}
\begin{bmatrix} \omega_x \\ \omega_y \\ \omega_z\end{bmatrix}
\end{equation}

The equation describing the evolution of translational motion of the quadrotor as derived in \cite{martin2009} is of special interest to the estimator design that will be presented in following sections.

\begin{align}\label{eq_force}
m\mathbf{\dot{V}} &= m\textbf{g} - k_T \sum_{i=1}^4 \omega_i^2 \mathbf{b_3} -\lambda_1\sum_{i=1}^4 \omega_i \tilde{\textbf{V}}
\end{align}

where 

\begin{align*}
\textbf{V} &= \textup{Velocity of $\{B\}$ as observed from an inertial frame} \\
\textbf{g} &= \textup{gravity vector}\\
k_T &= \textup{thrust coefficient of propellers}\\
\lambda_1 &= \textup{a positive coefficient known as rotor drag coefficient}\\
\omega_i &= \textup{rotational velocity of $i^{th}$ rotor, $i\in \{1,2,3,4\}$ }\\
\tilde{\textbf{V}} &= \textup{projection of $\textbf{V}$ on to the propeller plane}\\
m &=\textup{mass of the quadrotor}
\end{align*}

\begin{figure}[htb]
\begin{center}
\includegraphics[width=0.5\textwidth]{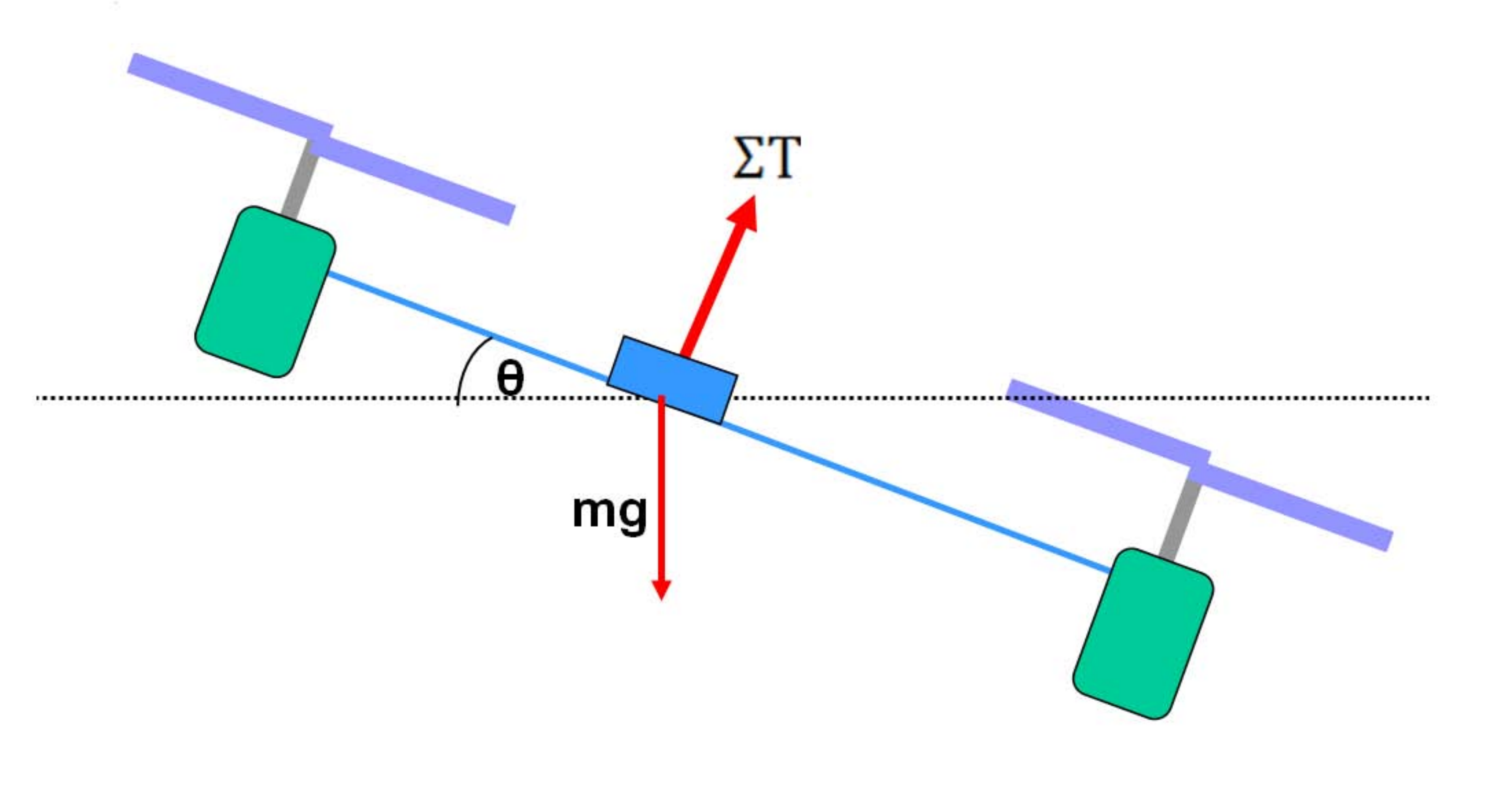}
\end{center}
\caption{Schematic of a quadrotor immediately after tilting sideways, but before it starts moving. $\Sigma T$ is the summation of propeller thrusts and corresponds to the second term in (\ref{eq_force})}
\label{quad_fig_a}
\end{figure}

Equation (\ref{eq_force}) sheds light on two key aspects of the quadrotor. First and the most obvious is the fact that the thrust force is perpendicular to the propeller plane, and thus has no effect on motion along that plane. Secondly and more importantly, we see the presence of a force which is proportional to the translational velocity of the quadrotor. For an intuitive description of this force, we refer readers to Fig. \ref{quad_fig_a} which shows a cross section of a quadrotor in flight, and provide below a simplified explanation of the origin of this force.

Fig. \ref{quad_fig_a} shows a quadrotor in a hypothetical state where it has tilted sideways to initiate a translation in a horizontal direction but immediately before it gains any translational motion. At this point thrust from propellers and gravity are the only forces acting on our simplified quadrotor model. As is obvious, in this particular state, thrust force generated from the propellers is perpendicular to the propeller plane.

\begin{figure}[htb]
\begin{center}
\includegraphics[width=0.5\textwidth]{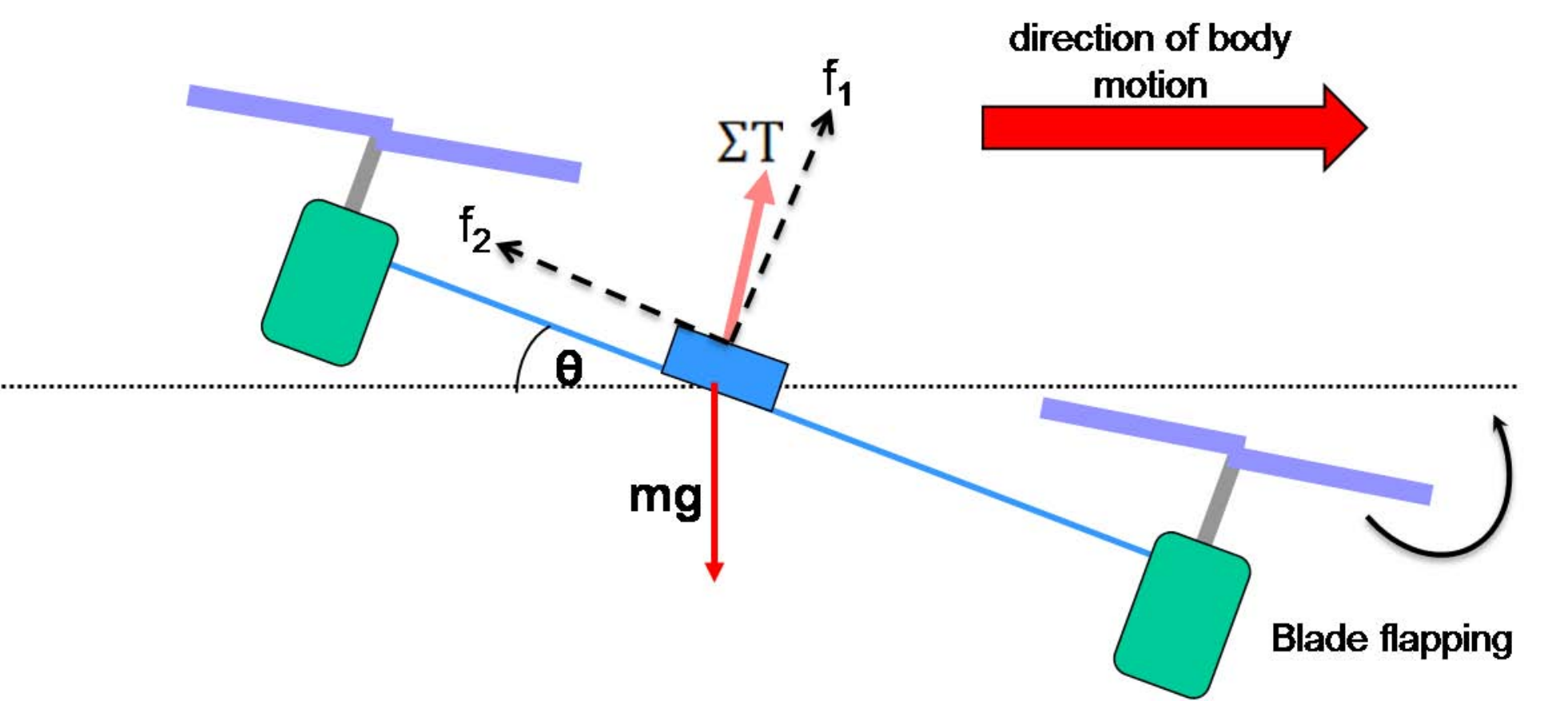}
\end{center}
\caption{Quadrotor, after tilting, starts moving sideways. $f_1$ and $f_2$ are the orthogonal components of $\Sigma T$}
\label{quad_fig_b}
\end{figure}

\begin{figure}[htb]
\begin{center}
\includegraphics[width=0.5\textwidth]{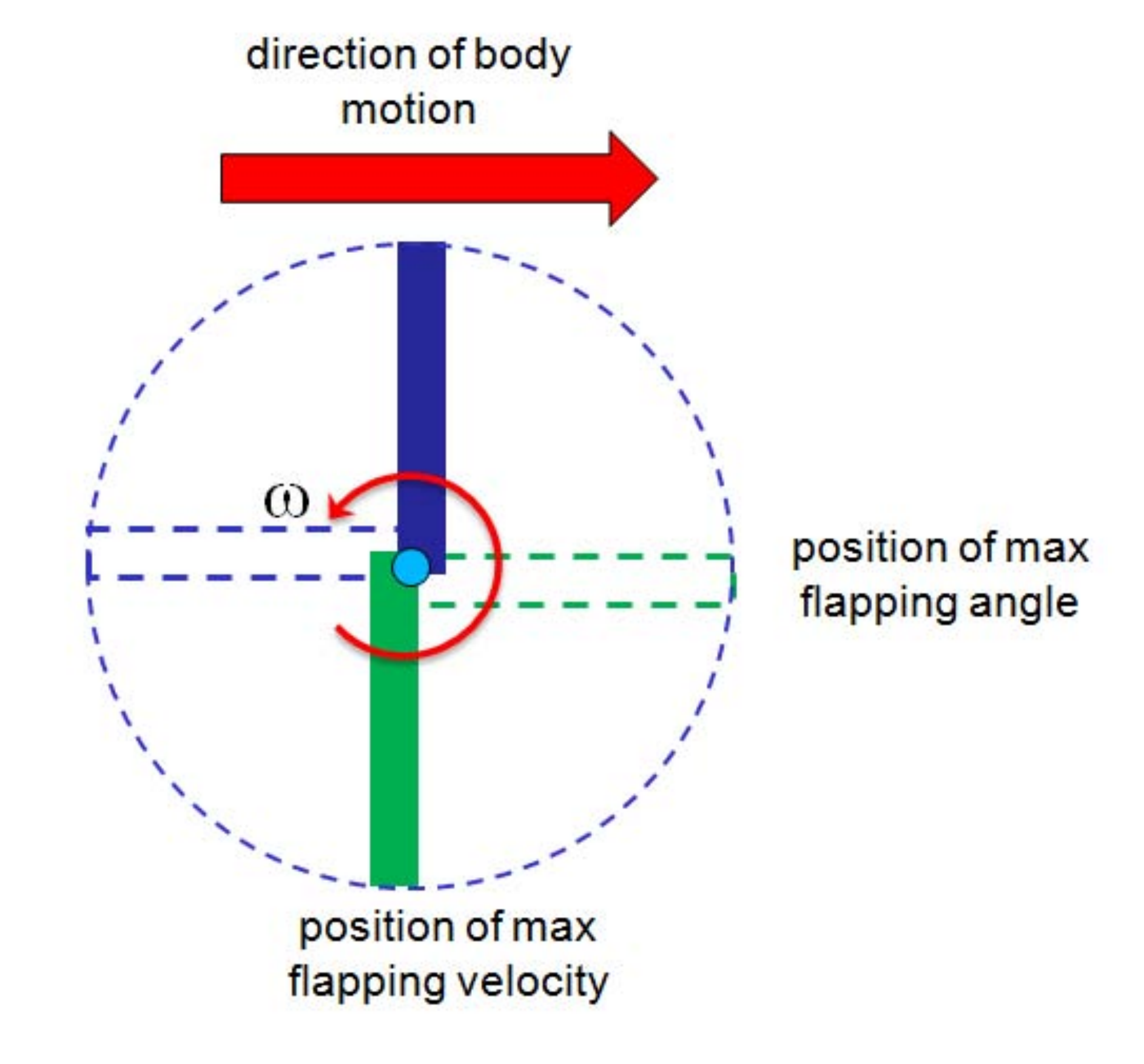}
\end{center}
\caption{As the propeller blades rotate, flapping is determined by their position with respect to the direction of motion of the propeller as a whole.}
\label{quad_fig_c}
\end{figure}

As stated, the state depicted in Fig. \ref{quad_fig_a} is hypothetical in the sense that even the slightest tilt of the quadrotor will induce translational motion. Fig. \ref{quad_fig_b} shows a more realistic situation in which quadrotor now moves right with a non-zero velocity. For a propeller with two blades, we can now identify a retreating and an advancing blade, as shown in blue and green respectively in Fig. \ref{quad_fig_c}. The velocity of the advancing blade with respect to free air is higher than that of the retreating blade, due to the translational velocity of the whole quadrotor. This creates a force imbalance between the two blades of the same propeller and thus causes the blades to flap up and down as they rotate. Blade flapping forces the propeller to rotate out of plane and the flapping angle of a blade is at a maximum just before it transitions from advancing state to retreating state or vice versa. As shown in Fig. \ref{quad_fig_b}, blade flapping causes the thrust force of the propeller to be tilted in a direction which opposes the motion of the quadrotor. As the amount of blade flapping is dependent on the translational velocity of the quadrotor, the component of thrust force along the body plane is also a function of that velocity. The last term in (\ref{eq_force}) models the impact of this component of the thrust force on the translational motion of the quadrotor. If one is to place an accelerometer on-board the quadrotor, with its' sensing axis parallel to the propeller plane, that accelerometer will measure a force that is roughly proportional to the velocity of the quadrotor along the same axis. In fact, in the next section, it is shown that this is the only significant force that the said accelerometer will sense. (Interestingly, (\ref{eq_force}) ignores the aerodynamic drag experienced by a body moving through air, which is usually a function of the square of the velocity. This can be justified for quadrotor MAVs that move at relatively low speeds.) This is \textit{the} unique characteristic of quadrotor MAVs that will later be exploited to the benefit of the state estimator.

To conclude this section, we re-write (\ref{eq_force}) using $^b{\textbf{V}}$ (i.e.\ \textbf{V} in $\{B\}$ frame) to facilitate the estimator design. After  neglecting the second order terms that appear due to coordinate frame transformation, the first two components of $^b\dot{\textbf{V}} \in \{^b\dot{v}_x , ^b\dot{v}_y , ^b\dot{v}_z\}$ can be written as

\begin{equation}
\left. \begin{aligned}\label{eq_vels}
^b\dot{v}_x &\approx -g\sin \theta - \frac{k_1}{m}\, ^bv_x  \\
^b\dot{v}_y &\approx g\cos \theta \sin \phi - \frac{k_1}{m}\, ^bv_y 
\end{aligned}
\right \}
\end{equation}

where 
\begin{align}
k_1 &= \lambda_1\sum_{i=1}^4 \omega_i \nonumber
\end{align}

In what follows, we assume that $k_1$ is a positive constant considering the fact that the summation of propeller rotational rates are fairly constant during smooth flight.

\section*{Inertial Sensors in Quadrotors}
This article is concerned with the quadrotor state estimators based on inertial sensors and specifically with accelerometers and gyroscopes. For simplicity, we assume that a triad of accelerometers and gyroscopes are mounted at the centre of mass of the quadrotor body. For both types of sensors we adhere to standard MEMS error models \cite{park2004}.

Gyroscopes measure the  instantaneous rotational rate of the body  with respect to the inertial frame, and their measurements can be modelled independently of the equations of motion of the moving platform to which they are attached.

\begin{align}
g_i &= \Omega_i + \beta_{gi} + w_{gi} \label{eq_gnoise}\\
\dot{\beta}_{gi} &= -\frac{1}{\tau_{gi}}\beta_{gi} + w_{\beta gi}      \label{eq_gbias}   
\end{align}

where $\beta_{gi}$ is the bias of $i^{th}$ gyroscope and $\tau_{gi}$ is the time constant of $i^{th}$ gyroscope bias. $w_{gi}$ and $w_{\beta gi}$ are zero mean White Gaussian Noise (WGN) terms.

In contrast, accelerometers measure a combination of inertial and gravitational acceleration, and their measurements can be expressed using the equations of motion governing  the body they are mounted on. Perhaps one of the best example of the value of this strategy is the case of a triad of accelerometers mounted on a quadrotor platform. Denoting by $\tilde{a}_i$ the acceleration that would be measured by an ideal accelerometer, we combine the accelerometer measurement model with (\ref{eq_force}) to arrive at:

\begin{align}
\tilde{\textbf{a}} &= \mathbf{\dot{V}} - \textbf{g}  \nonumber\\
&= - k_T \sum_{i=1}^4 \omega_i^2 \mathbf{b_3} -\lambda_1\sum_{i=1}^4 \omega_i \tilde{\textbf{V}}\label{eq_acc}
\end{align}  

Equation (\ref{eq_acc}) describing the readings obtained from an on-board triad of accelerometer is unique to quadrotors and is of critical importance to a state estimator in that context. As stated in the previous section, equation (\ref{eq_acc}) shows that the accelerometers along $\mathbf{b_1}$ and $\mathbf{b_2}$ coordinate axes are only sensitive to a force which is dependant on the projection of the quadrotor translational velocity on to $\mathbf{b_1}$, $\mathbf{b_2}$ plane. Furthermore, the component of the gravitational acceleration in the body frame (which is typically large compared to inertial accelerations of slow moving vehicles) no longer influences the accelerometer measurement. In the following section, we will exploit this unique property to design a better state estimator for quadrotors. 

\section*{Estimator Design}
The goal here is to design a state estimator for the quadrotor, giving due regard to the dynamic and kinematic equations presented in the previous sections. For this, we propose a six state, Extended Kalman Filter (EKF) based state estimator. The filter states are:

\begin{align*}
\phi &- \textup{Roll angle in current orientation estimate}\\
\theta &- \textup{Pitch angle in current orientation estimate}\\
\beta_{gx} &- \textup{Bias in X axis gyroscope}\\
\beta_{gy} &- \textup{Bias in Y axis gyroscope}\\
^bv_x &- \textup{X velocity component of quadrotor in body frame}\\
^bv_y &- \textup{Y velocity component of quadrotor in body frame}
\end{align*}

\subsection*{Process Model} 
Equations (\ref{eq_rotkin}),  (\ref{eq_vels}) - (\ref{eq_gbias}) form the EKF process equation. Out of the three Euler angles we can only estimate $\phi$ and $\theta$ as the equations are expressed in a form independent of the yaw angle  $\psi$.

\begin{equation}
\left.
\begin{aligned}\label{eq_rotkin2}
\dot{\phi} &= (g_x - \beta_{gx} + w_{gx}) + \tan\theta \cos\phi(g_z - \beta_{gz})  \\ & \qquad + \tan\theta \sin\phi(g_y - \beta_{gy} + w_{gy}) \\
\dot{\theta} &= \cos\phi (g_y - \beta_{gy} + w_{gy}) - \sin\phi(g_z - \beta_{gz})\\
\end{aligned}
\right \}
\end{equation}

\vspace{2mm}
\begin{equation}
\left.
\begin{aligned}\label{eq_bias}
\dot{\beta}_{gx} &= -\frac{1}{\tau_{gx}}\beta_{gx} +w_{\beta gx} \\
\dot{\beta}_{gy} &= -\frac{1}{\tau_{gy}}\beta_{gy} +w_{\beta gy} \\
\end{aligned}
\right \}
\end{equation}

\vspace{2mm}
\begin{equation}
\left.
\begin{aligned}\label{eq_vel}
^b\dot{v}_x &= -g\sin \theta - \frac{k_1}{m}\, ^bv_x   + w_{\alpha x} \\
^b\dot{v}_y &= g\cos \theta \sin \phi - \frac{k_1}{m}\, ^bv_y + w_{\alpha y}\\
\end{aligned}
\right \}
\end{equation}

where, $w_{\alpha x}$ and $w_{\alpha y}$ are WGN terms included to account for the model imperfections in (\ref{eq_vels}).

Equations (\ref{eq_rotkin2}) , (\ref{eq_bias}) and (\ref{eq_vel}) together describe the process dynamics of the estimator. The resulting system can be represented as a non-linear function of states, control inputs and noise terms.

\begin{align*}
\dot{\textbf{x}} = f(\textbf{x}, \textbf{u},\textbf{w})
\end{align*}

\subsection*{Measurement Model}
Observations of the EKF are the measurements from X and Y accelerometers, which are aligned  respectively with $b_1$ and $b_2$. Measurement equations can be easily derived from (\ref{eq_acc}), after including accelerometer noise terms, which are assumed to be Gaussian.

\begin{equation}
\left.\begin{aligned}\label{eq_ms}
a_x &=  - \frac{k_1}{m}\, ^bv_x + w_{ax} \\
a_y &=  - \frac{k_1}{m}\, ^bv_y + w_{ay}  \\
\end{aligned}
\right\}
\end{equation}

where $a_x$ and $a_y$ are respectively the measurements from the X and Y axis accelerometers on-board the quadrotor. Here we assume that accelerometer biases are random constant values which can be compensated for, offline.

\subsection*{EKF Mechanization Equations}

For the mechanization of the Extended Kalman Filter, the discrete state transition matrix $A_k$ should be calculated. For this we first  calculate $F$, which is the Jacobian matrix of partial derivatives of $f$ with respect to $\textbf{x}$. Then $A_k$ is calculated by discretization of the Jacobian matrix.

\begin{align}
F(t) &= \frac{\partial{f(\textbf{x}, \textbf{u},\textbf{w})}}{\partial{\textbf{x}}} \Big|_{\hat{\textbf{x}}_k,\textbf{u}_k} \nonumber
\end{align}

Discretization is performed with a truncated Taylor series approximation and a sample time of $T_s$, resulting in, 

\begin{align}
A_k &= I + F(t)T_s \nonumber
\end{align}

In deriving the discrete process noise matrix $Q_k$, we assume that noise terms in (\ref{eq_rotkin2}) and (\ref{eq_bias}) are uncorrelated with each other as well as with accelerometer noise terms.

\begin{align}
\textbf{w} &= \begin{bmatrix} w_{gx} & w_{gy} & w_{\beta gx} & w_{\beta gy} & w_{\alpha x} & w_{\alpha y} \end{bmatrix}^T \nonumber \\
W(t) &= diag \begin{bmatrix} \sigma^2_{gx} & \sigma^2_{gy} & \sigma^2_{\beta gx} & \sigma^2_{\beta gy} & \sigma^2_{\alpha x} & \sigma^2_{\alpha y} 
\end{bmatrix} \nonumber\\ 
Q(t) &= G(t) W(t) G^T(t) \nonumber
\end{align}

The first four terms of the $(t)$ are the noise variances of gyroscope sensors and their biases. These can be found by experimentation with actual sensors. Last two terms, which correspond to the uncertainty in (\ref{eq_vel}) were approximated first and then fine tuned for optimum performance of the estimator. Also,
\begin{align}
G(t)&=\frac{\partial{f(\textbf{x}, \textbf{u},\textbf{w})}}{\partial{\textbf{w}}} \Big|_{\hat{\textbf{x}}_k,\textbf{u}_k} \nonumber
\end{align}

 Discretization of $Q(t)$ results in $Q_k$.

\begin{align}
Q_k &= \int _0^{T_s} AQ(\tau)A^T \, d\tau \nonumber
\end{align}

Measurement matrix $H$ required for the EKF can be directly obtained from (\ref{eq_ms}) as,

\begin{align}
H = \begin{bmatrix} 0 & 0 & 0 & 0 & -k_1/m & 0 \\
						  0 & 0 & 0 & 0 & 0 & -k_1/m 
\end{bmatrix} \nonumber
\end{align}

Assuming uncorrelated errors in accelerometer measurements, measurement noise matrix $R_k$ becomes diagonal, consisting only of the noise variances of the $X$ and $Y$ accelerometers.

\begin{equation}
R_k = diag \begin{bmatrix} \sigma^2_{ax} & \sigma^2_{ay} \end{bmatrix} \nonumber
\end{equation}

For initialisation, all states of the filter are set to zero and their error covariances are set to small positive values reflecting the uncertainty in initial estimate. With multiple experimental runs, it was found that  changes of up to 100\% in the initial values and the noise variances have negligible effect on filter performance. We attribute this robustness of the estimator to the linear measurement model and not-so-strong non-linearities in the process equations.

EKF state prediction was carried out with the use of a $2^{nd}$ order Runga-Kutta integrator. Covariance projection, Kalman gain calculation, state update and covariance update equations of the estimator take their standard forms as detailed in \cite{grewal2001}.

\section*{ARDrone Quadrotor and the experiments}
\begin{figure}[htb]
\begin{center}
\includegraphics[width=0.4\textwidth]{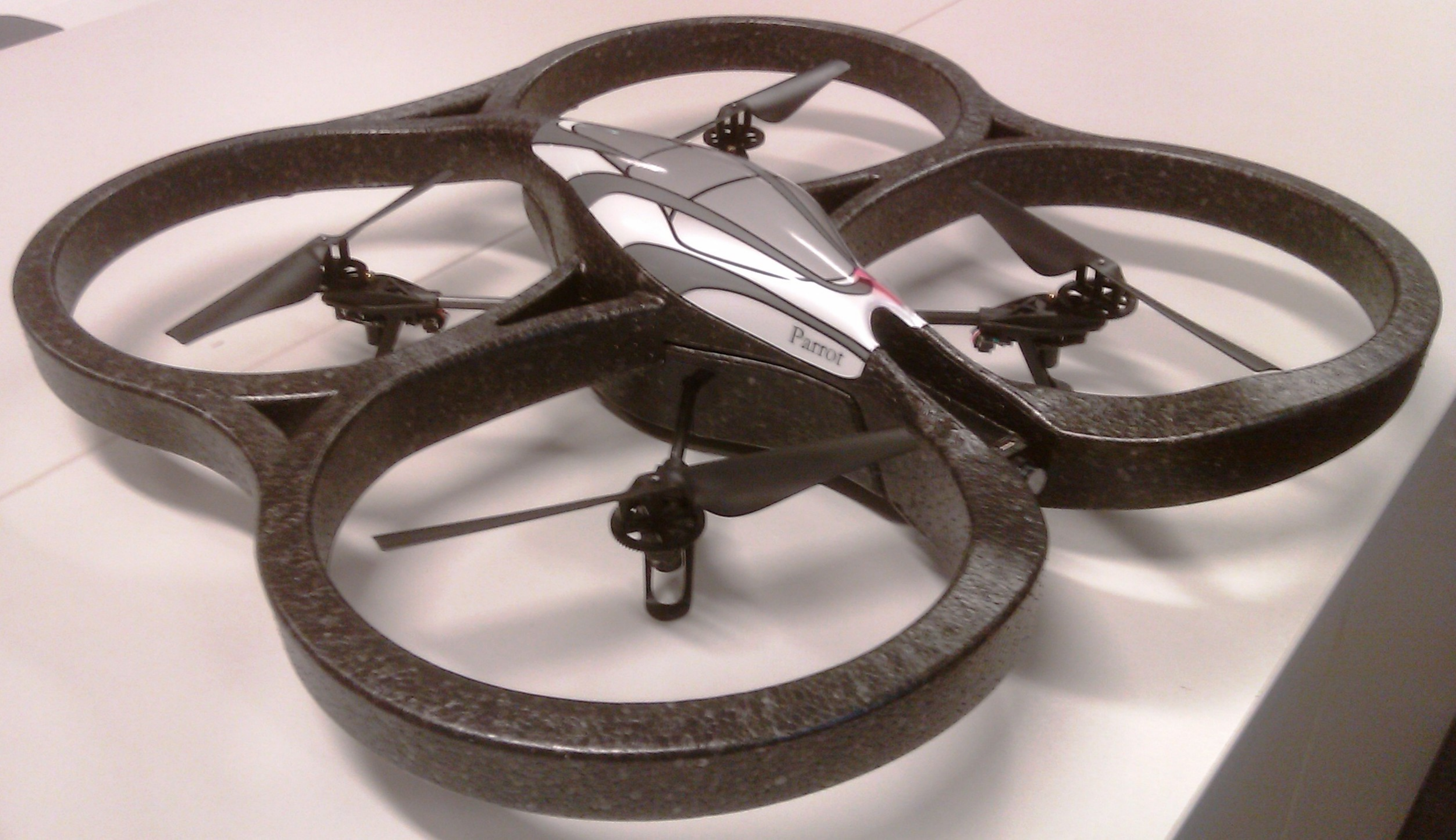}
\end{center}
\caption{ARDrone Quadrotor used for experiments}
\label{fig_drone}
\end{figure}

The quadrotor platform used for the experiments presented in this article is the Parrot ARDrone \cite{ardrone2012} (see Fig. \ref{fig_drone}) ARDrone weighs about 420g including the protective hull and has a flight time of about 10 minutes. Straight out of the box, ARDrone is an extremely stable quadrotor platform and therefore is an excellent platform for quadrotor based research. It is equipped with a wide array of sensors including triad of accelerometers, triad of gyroscopes, two cameras -one  facing front and other facing down- and downward pointing sonar sensors.  All sensor data from the ARDrone are wirelessly transmitted to a ground station PC either running Windows or Linux. An open source C API is provided which can be easily extended to develop application on the ground station to process incoming sensor data and to send out control commands to the ARDrone. It is also equipped with a pre-programmed closed source attitude control system, which takes care of the low-level stabilisation and control tasks, while providing users the ability to develop applications for higher level navigational tasks. 

It is desirable to have ``ground truth" states trajectories for performance evaluation of the proposed estimator. Therefore, all our ARDrone experiments were performed in a Vicon motion capture environment. The Vicon motion capture system uses a set of reflective markers rigidly attached to the quadrotor body, which are observed by 8 fixed IR cameras to directly compute the attitude and position of the quadrotor with respect to the Vicon coordinate frame. 

In a typical experiment, ARDrone was manually piloted within the Vicon environment (approximately 6 $\times$ 4 $\times$ 3 m) using a joy stick attached to the ground station computer. The inertial sensor data were continuously streamed to the ground station computer at 200Hz and were stored for post processing. Vicon generated state estimated were also stored in a separate PC. Matlab computing environment was used for post processing of both inertial and Vicon data.

A critical parameter that needs to be precomputed for the estimator is the rotor drag coefficient $\lambda_1$. Since a theoretical calculation of this parameter is a complex task, we resorted to an experimental estimation method. The basic methodology adopted here is to obtain the accelerometer measurements and ground truth velocity data of a few flight tests. A rough estimate of the parameter $k_1$ (which incorporates $\lambda_1$) can then be obtained by formulating (\ref{eq_ms}) as a least-squares problem. For the ARDrone, the best estimate for the parameter $k_1$ was found to be 0.57.  This parameter estimation task was run only once and the derived $k_1$ values was used for all subsequent estimation tasks.

\section*{Experimental Results} 
\begin{figure}[htb]
\begin{center}
\includegraphics[width=0.5\textwidth]{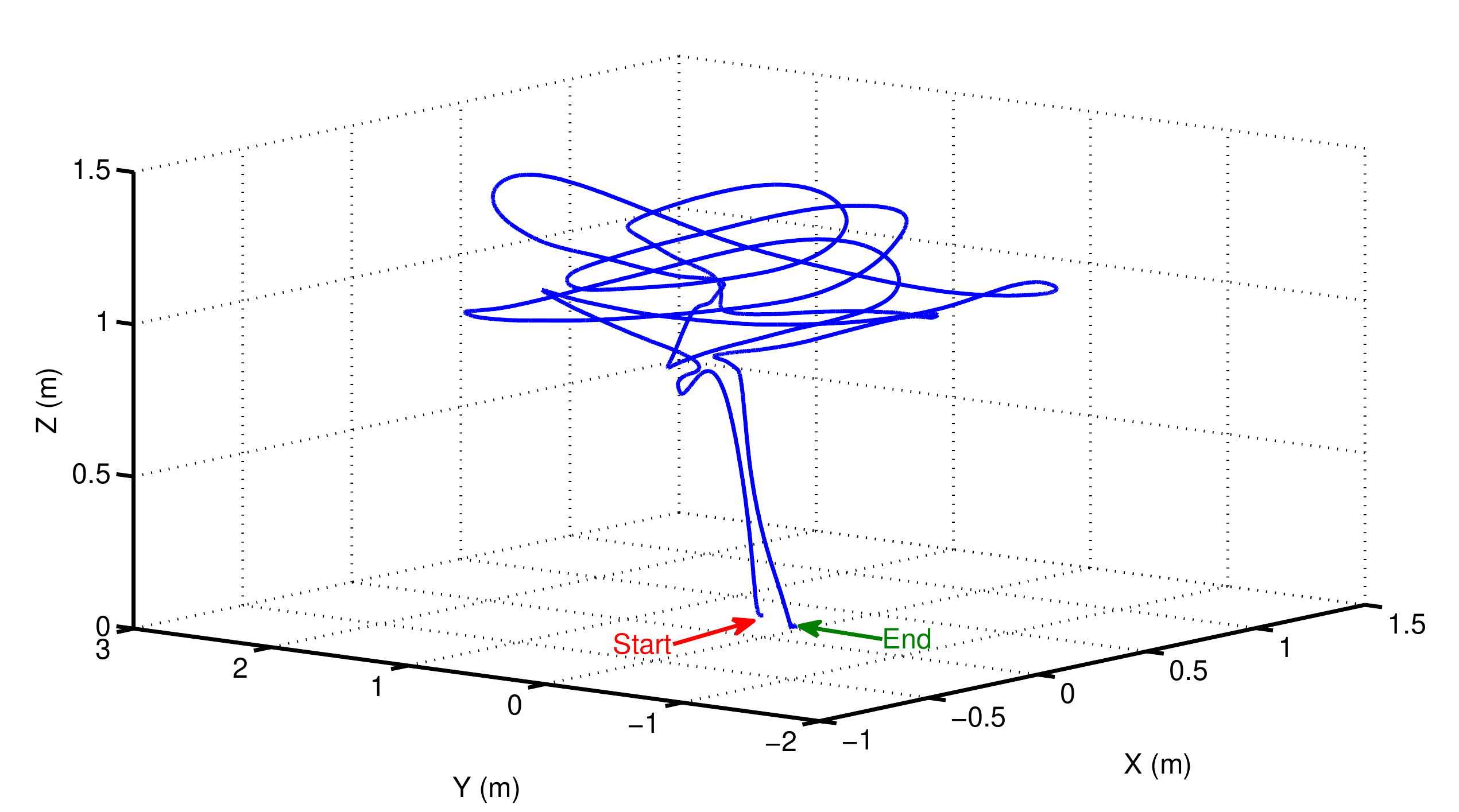}
\end{center}
\caption{Three-dimensional flight path of the ARDrone experiment }
\label{3d_flight_path}
\end{figure}

During one experiment, the AR Drone was manually operated within the Vicon environment, moving freely while keeping the height approximately constant.  A three-dimensional trace of the path taken by the MAV in a typical experiment is shown in Fig. \ref{3d_flight_path}. The results presented in the following sections are based on the data gathered from this experiment.

\begin{figure}[htb]
\centering
\subfigure[]{
\includegraphics[width=0.5\textwidth]{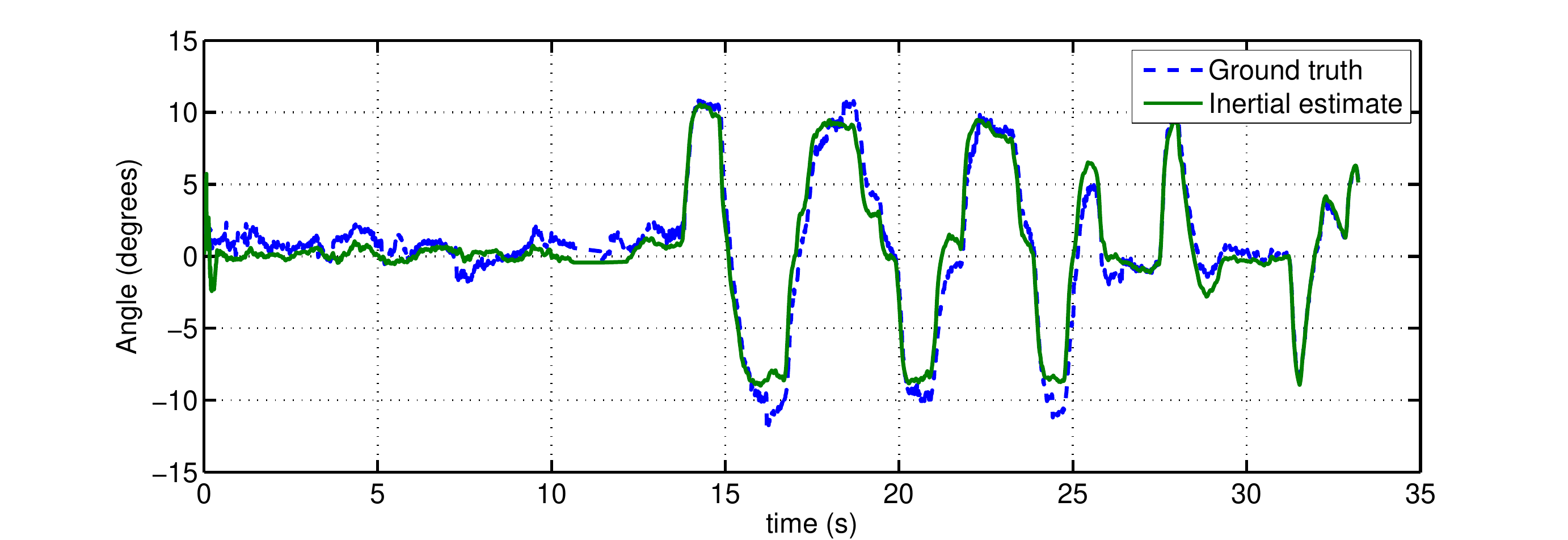}
\label{phi}
}
\subfigure[]{
\includegraphics[width=0.5\textwidth]{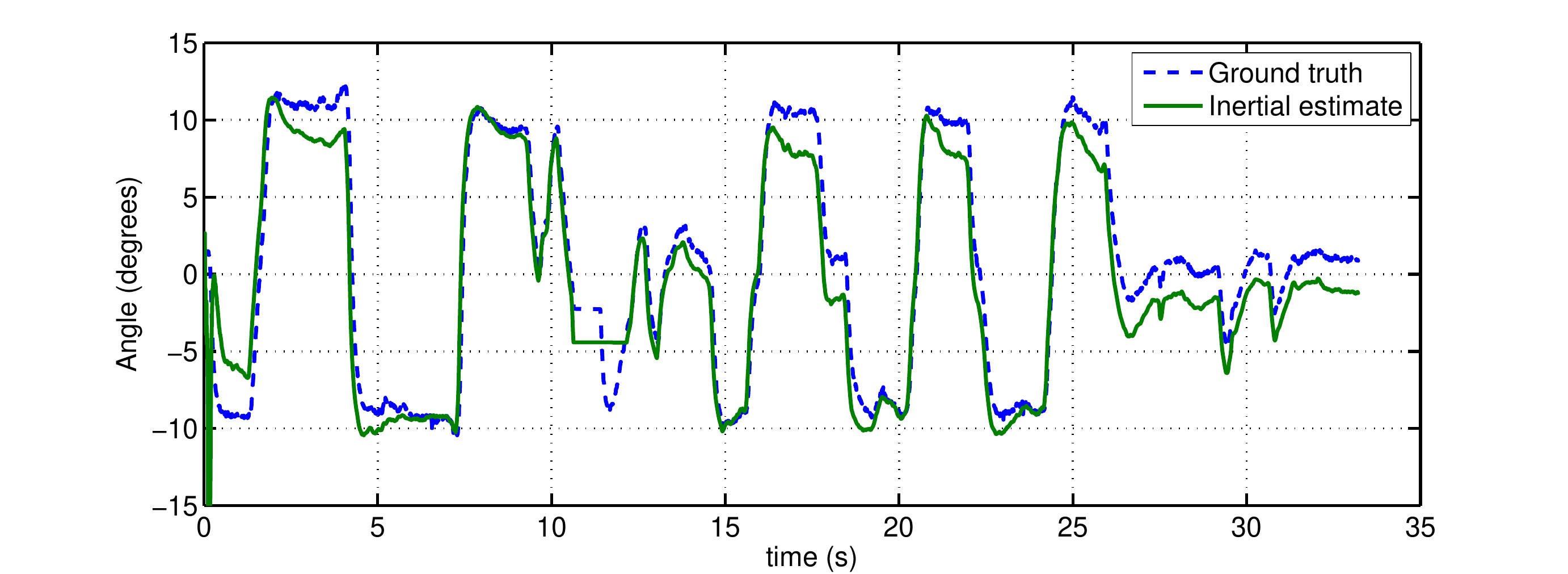}
\label{theta}
}
\caption[]{Comparison of ground truth and inertial attitude estimates of AR Drone. \subref{phi} Roll angle ($\phi$), \subref{theta} Pitch angle ($\theta$)}
\label{fig_st2_angles}
\end{figure}

\begin{figure}[htb]
\centering
\subfigure[]{
\includegraphics[width=0.5\textwidth]{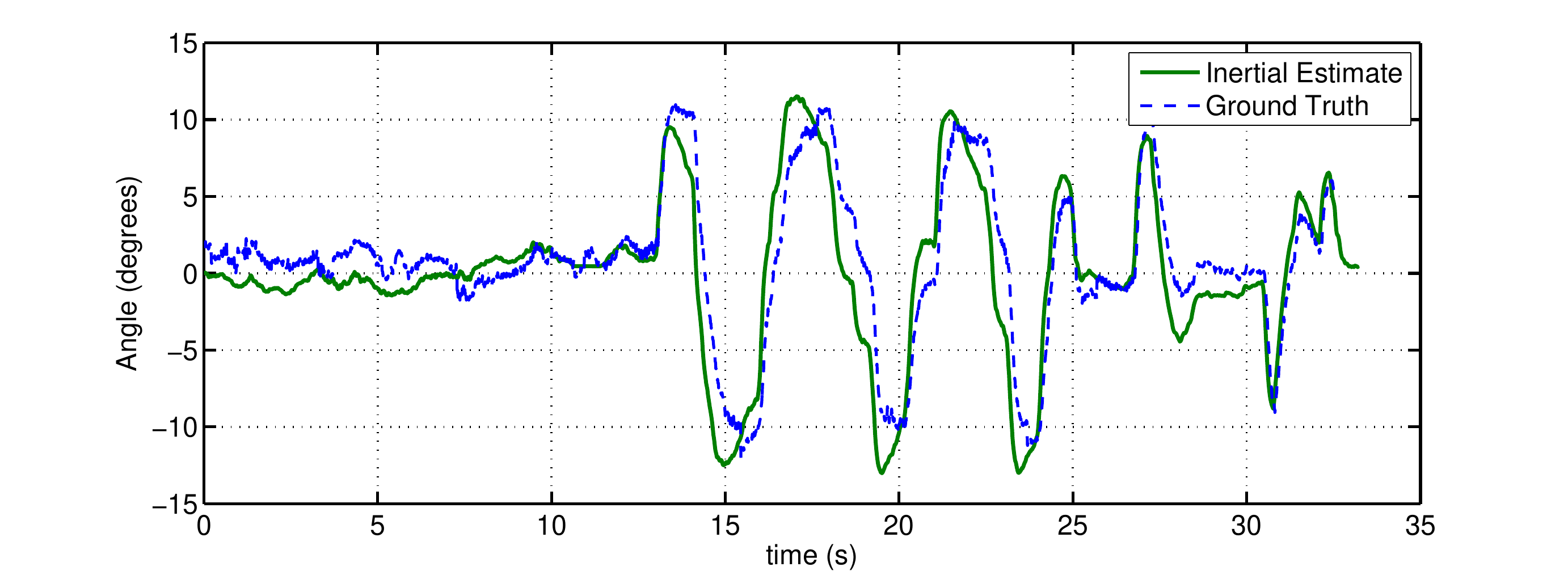}
\label{gen_phi}
}
\subfigure[]{
\includegraphics[width=0.5\textwidth]{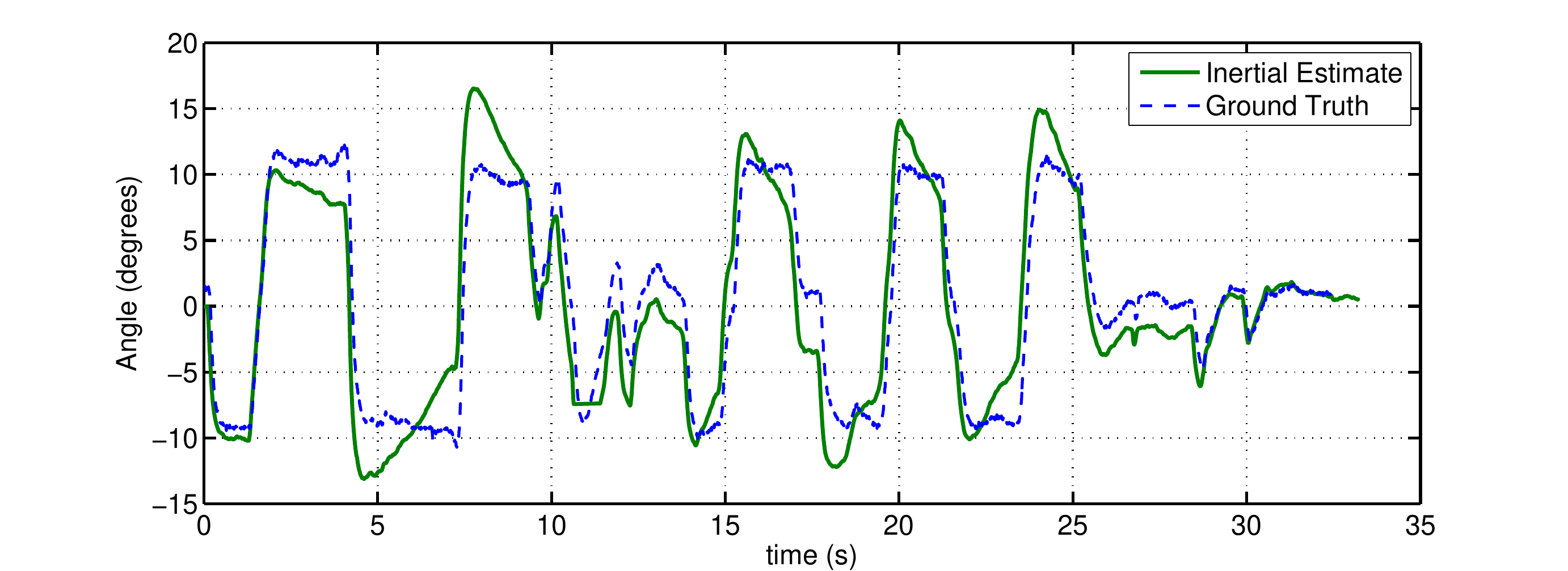}
\label{gen_theta}
}
\caption[]{Comparison of ground truth and inertial attitude estimates of AR Drone, obtained from the generic estimator. \subref{gen_phi} Roll angle ($\phi$), \subref{gen_theta} Pitch angle ($\theta$)}
\label{fig_gen_angles}
\end{figure}

Fig. \ref{fig_st2_angles} shows the attitude estimates  of the proposed EKF together with the ground truth obtained from the Vicon system. For comparison purposes, we have also plotted the attitude estimates from a generic estimator as detailed in \cite{dmw2010} in Fig. \ref{fig_gen_angles}. It is important to note the improvement in the pitch estimate of the proposed estimator over the generic estimator. This improvement is more pronounced in places where the quadrotor changes its flight direction (for example around 4.6 and 7.8 sec).  During those intervals, the quadrotor undergoes high inertial accelerations and the assumption that the accelerometer measurements are dominated by gravitational acceleration fails to hold. Thus generic attitude estimators based on this assumption produce erroneous  results. As expected the proposed EKF attitude estimates agrees more with the ground truth because such an assumption is not utilized in that design.  However, when the quadrotor is not undergoing considerable accelerations, the two attitude estimates converge and the generic estimator can perform just as well as the proposed method.

\begin{figure}[htb]
\centering
\subfigure[]{
\includegraphics[width=0.5\textwidth]{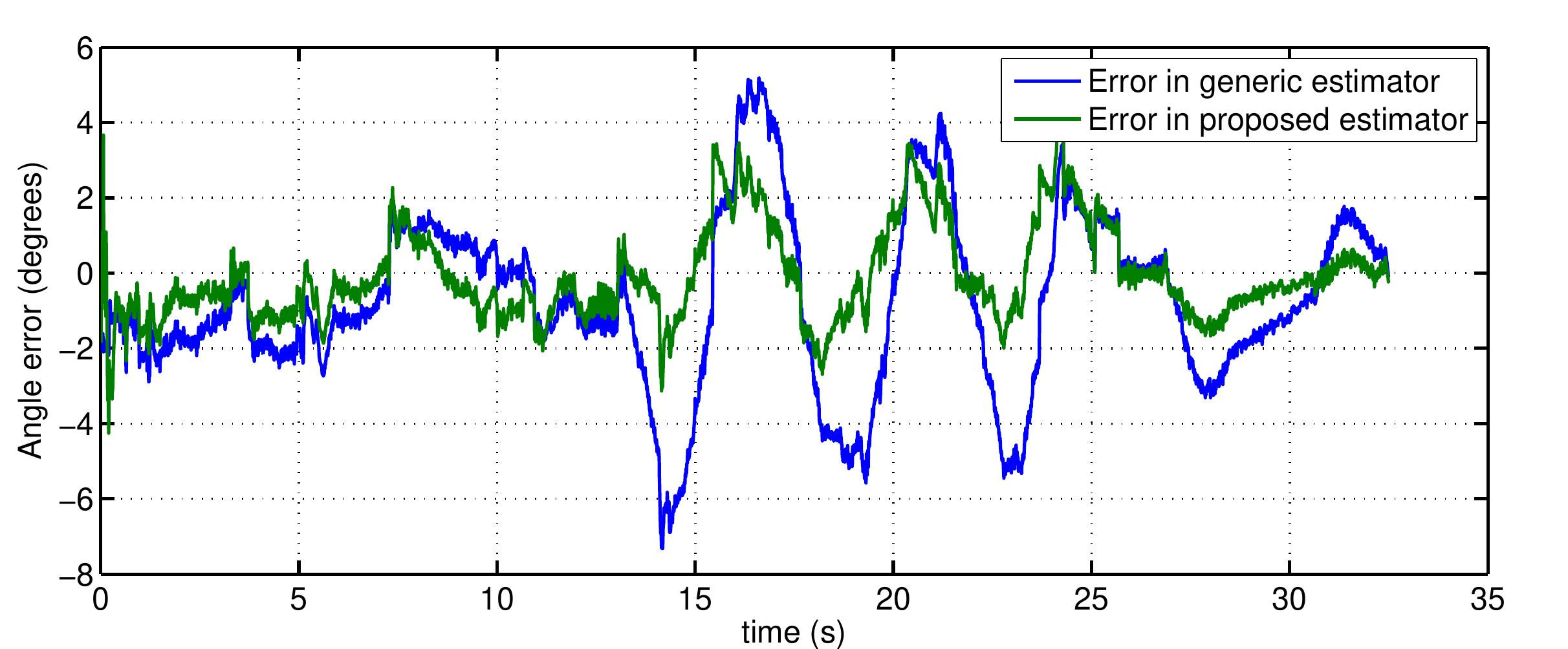}
\label{phi_error_comp}
}
\subfigure[]{
\includegraphics[width=0.5\textwidth]{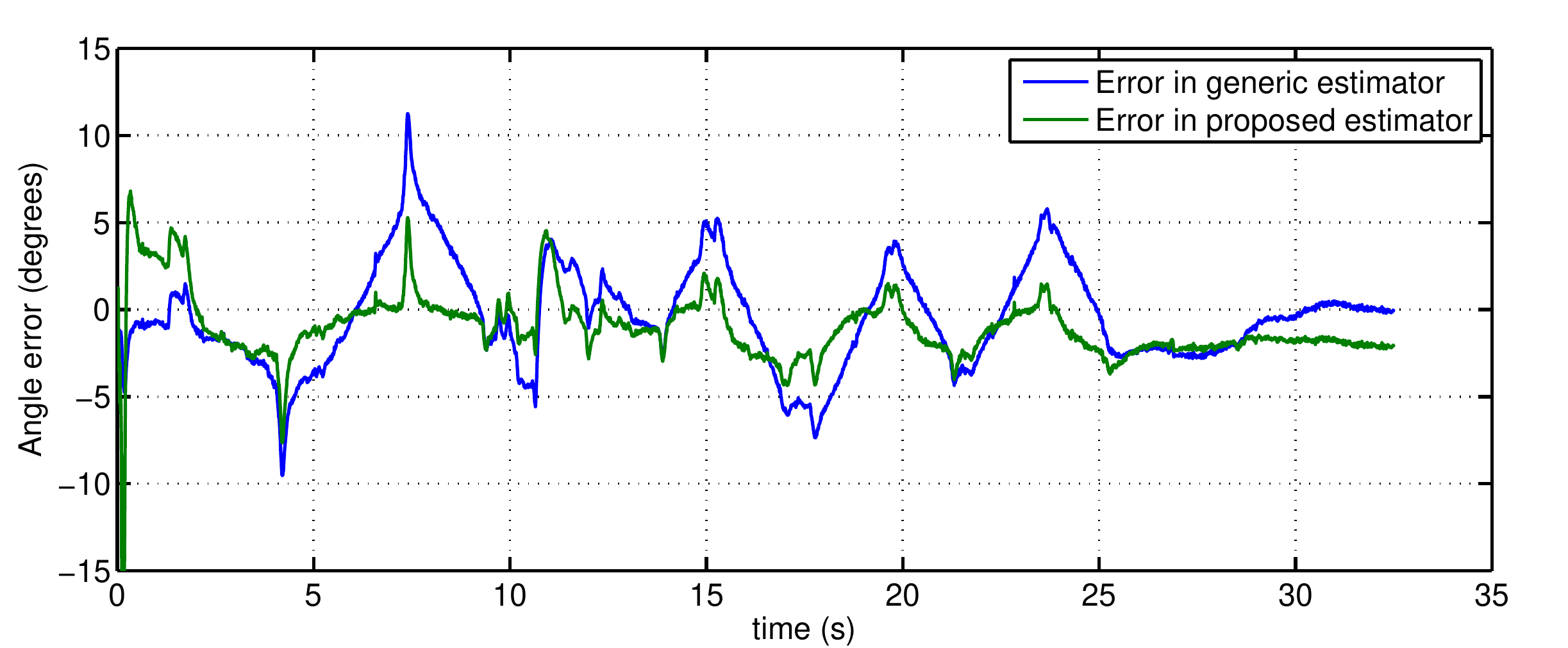}
\label{theta_error_comp}
}
\subfigure[]{
\includegraphics[width=0.5\textwidth]{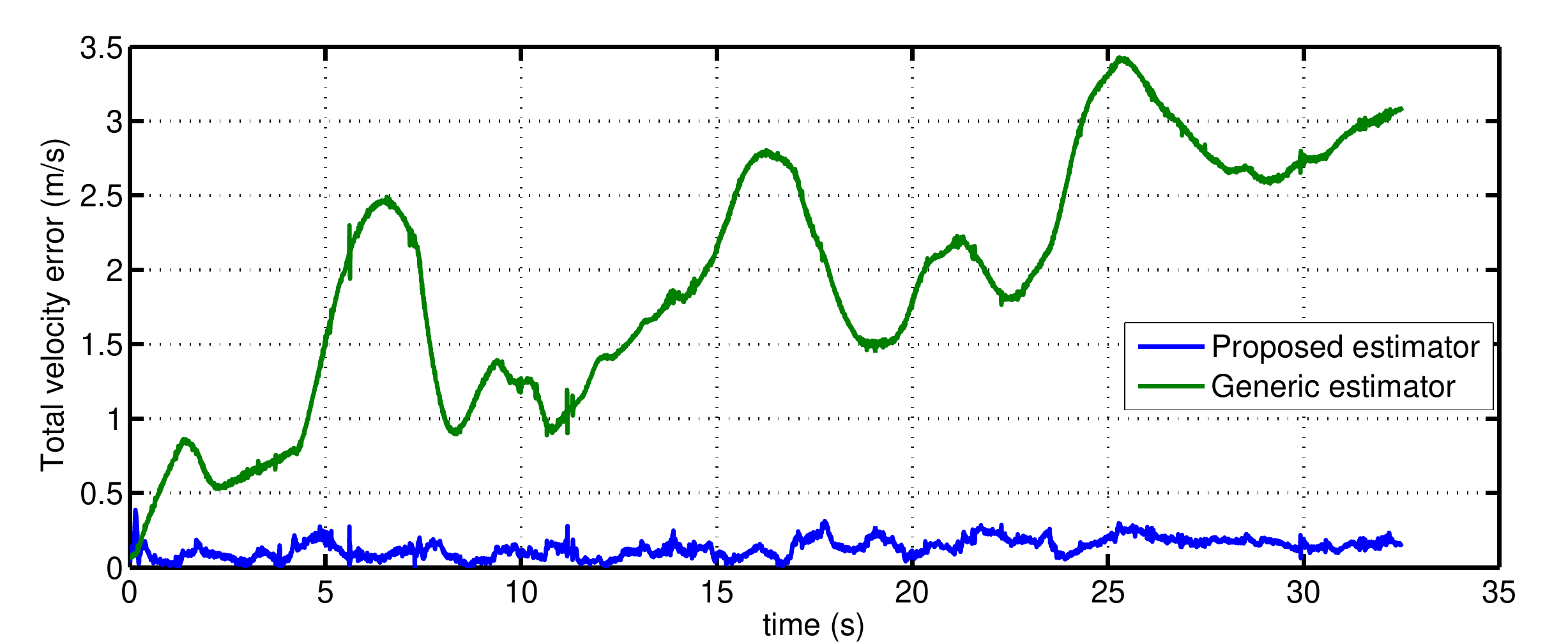}
\label{total_vel_err}
}
\caption[]{Estimation errors of both estimator designs. \subref{phi_error_comp} Roll angle ($\phi$) estimation error, \subref{theta_error_comp} Pitch angle ($\theta$) estimation error, \subref{total_vel_err}  Total velocity estimation error}
\label{fig_error_comp}
\end{figure}

Fig. \ref{phi_error_comp} and \ref{theta_error_comp} present a comparison between the errors in the roll and pitch attitude estimates of both the proposed EKF and the generic estimator. Even with the proposed EKF, unmodelled dynamics (such as displacement of accelerometer from the centre of mass of the quadrotor) causes an increase in estimation error when the quadrotor undergoes large accelerations. But overall, it is clear that the errors in the proposed design are considerably less than those of the generic design.

\begin{figure}[htb]
\centering
\subfigure[]{
\includegraphics[width=0.5\textwidth]{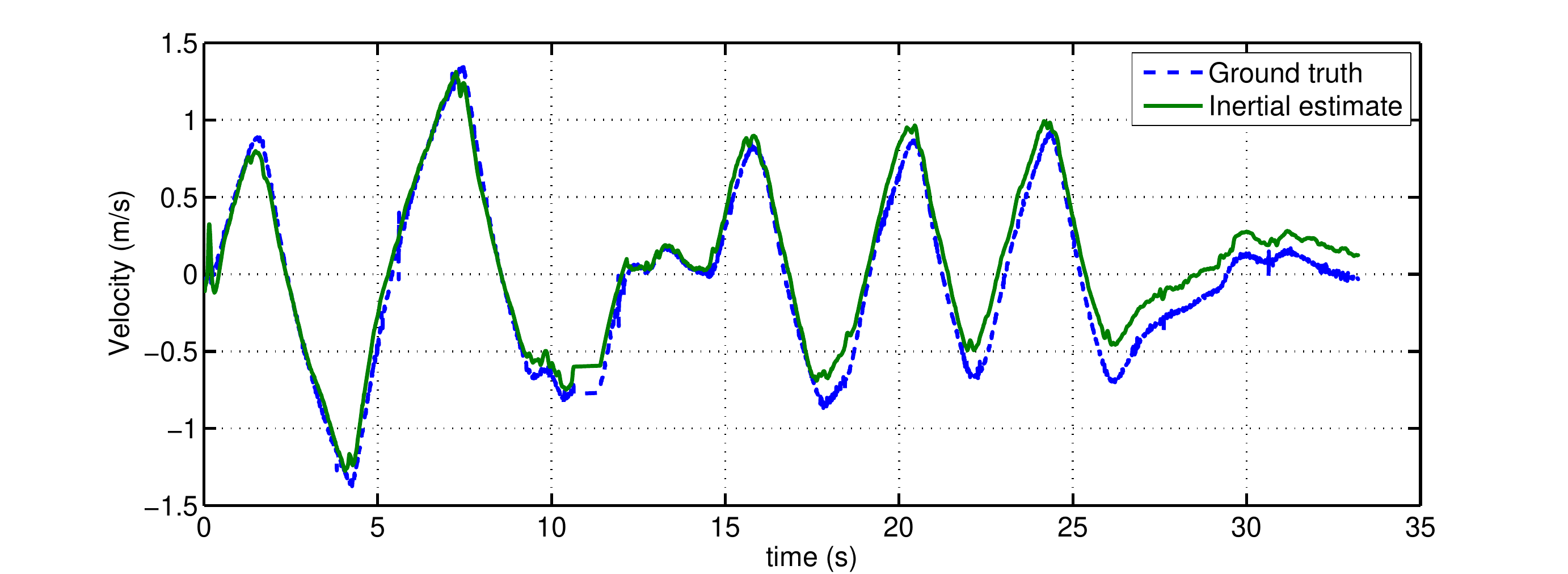}
\label{x_velocity}
}
\subfigure[]{
\includegraphics[width=0.5\textwidth]{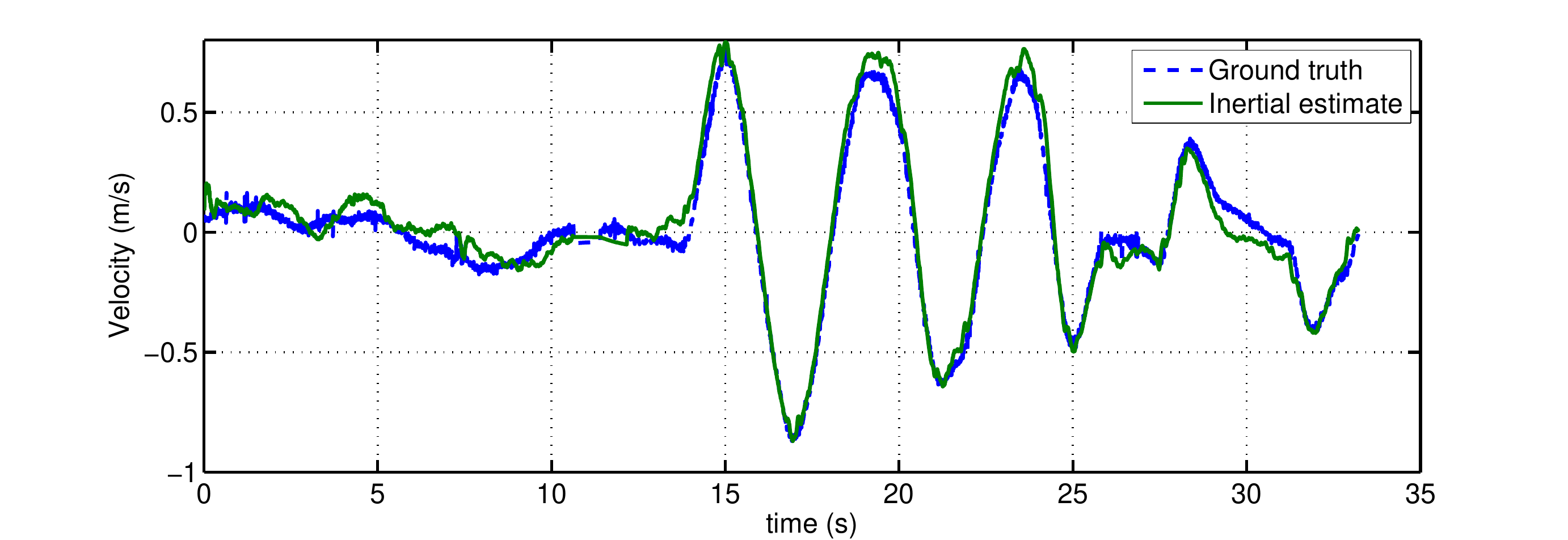}
\label{y_velocity}
}
\caption[]{Comparison of ground truth and inertial velocity estimates of AR Drone. \subref{x_velocity} X Velocity ($V_x$), \subref{y_velocity} Y Velocity ($V_y$)}
\label{fig_st2_vels}
\end{figure}

\begin{figure}[htb]
\centering
\subfigure[]{
\includegraphics[width=0.5\textwidth]{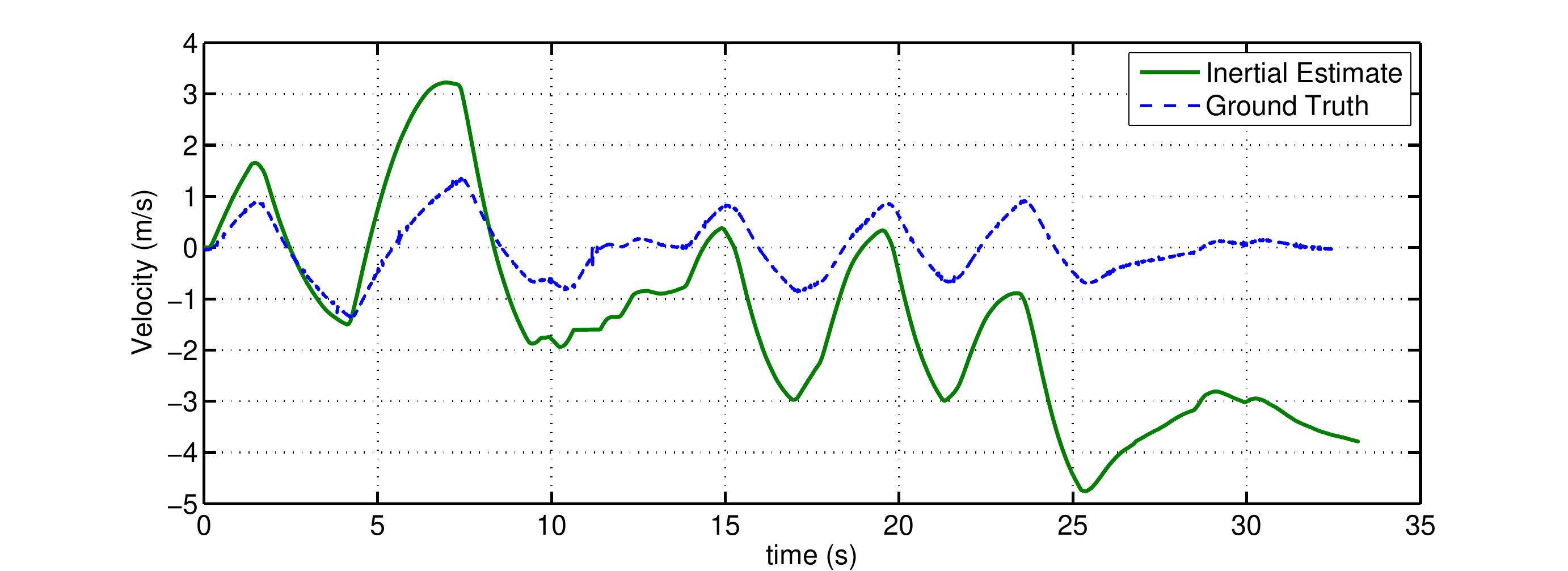}
\label{gen_x_velocity}
}
\subfigure[]{
\includegraphics[width=0.5\textwidth]{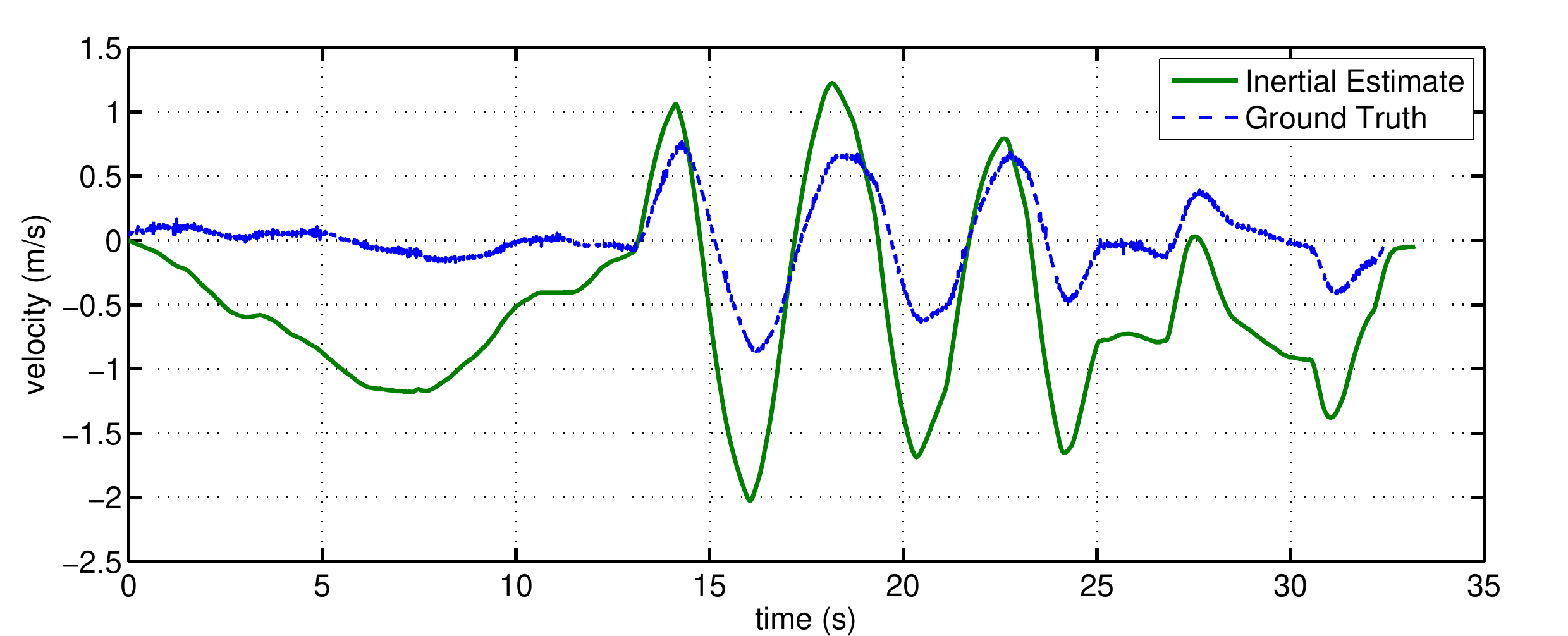}
\label{gen_y_velocity}
}
\caption[]{Comparison of ground truth and inertial velocity estimates of AR Drone, obtained from the generic estimator. \subref{gen_x_velocity} X Velocity ($V_x$), \subref{gen_y_velocity} Y Velocity ($V_y$)}
\label{fig_gen_vels}
\end{figure}

Fig. \ref{fig_st2_vels}  presents the velocity estimate from the proposed EKF together with the ground truth.  Again for comparison, Fig. \ref{fig_gen_vels} shows the velocity estimates in a generic design where, velocity is estimated by integrating inertial accelerations calculated by compensating the accelerometer measurements for gravity. A comparison between the errors in velocity estimate obtained from the proposed estimator and the generic estimator is shown in Fig. \ref{total_vel_err}, where total velocity error is the sum of root square errors of both $X$ and $Y$ axes. What is important to note is that the proposed strategy produces velocity estimates in which errors do not grow with time, while estimating velocity through direct integration of accelerations  as implemented in the conventional design leads to a significant drift.  As zero velocity updates, that can be used to correct this behaviour in land vehicles, are no longer viable with an MAV without some deliberate control strategies, this points to a significant advantage of the estimator proposed in this article.

\section*{Conclusion}
In this article, we presented a novel state estimator for quadrotor MAVs, where clear improvements in estimates stemming from the incorporation of quadrotor specific dynamical constraints were demonstrated. Our design is based on an EKF and is capable of estimating both roll and pitch angles of the attitude in addition to $X$ and $Y$ components of the body frame translational velocities within a bounded error. This estimator is applied to inertial data gathered from real world flight experiments. The resulting attitude and velocity estimates obtained match closely with the ground truth and are drift free.  

Before concluding the discussion on the estimator performance, we note that our design by itself is not a perfect solution to the problem of quadrotor state estimation. We believe that two key improvements need to be made to our design. First, an online estimation of the parameter $\lambda_1$ and accelerometer biases will improve estimation accuracy and ease the filter design process. Secondly, the estimation $\psi$ angle and velocity $^bv_z$ will improve the autonomy of the quadrotor. Our current research focuses on these improvements. 

In addition, we also expect to fuse the inertial information with exteroceptive sensors such as cameras and GPS. The two cameras in the ARDrone makes it an ideal platform for visual Simultaneous Localisation And Mapping (SLAM). One key drawback in employing monocular SLAM for MAVs is the unavailability of odometry for scale recovery. Another more obscure problem is the alignment of camera with the MAV body frame. From a control theoretic perspective, orientation of the body frame is what matters and misalignment of camera and body frames can lead to poor control performance in a SLAM only MAV state estimator. Both these problems can be solved by tightly integrating the estimation algorithm presented here with a monocular SLAM algorithm. We believe this to be an exciting research avenue.

\section*{Acknowledgements}
This work is supported by the Centre for Autonomous Systems, University of Technology Sydney.

\bibliography{jreferences}
\bibliographystyle{IEEEtran}

\end{document}